# Using the YOLOv12 Model for Verifying the Correct Color Sequence of Wires in Network Cables (Patch Cords) on the Production Line

**Amin Doroodchi***
Computer Department, Islamic Azad University, Beyza Branch, Beyza, Iran

**Danial Soleymani**
R&D at Nedaye Sabz Company, Isfahan Branch, Iran

*Corresponding Author: email

## Abstract

In the production process of network cables, ensuring the correct color sequence of wire pairs inside the standard connector plays a critical role in the final performance of the cable, as any misplacement or color-ordering error can lead to defective products and impose significant costs. Traditional inspection methods based on visual examination through digital microscopes are typically time-consuming, tedious, and prone to human error. In this study, an intelligent system based on the twelfth version of the YOLO[1] object detection model was developed to identify the position and verify the correct color sequence of wires in patch cords. The dataset used consisted of 2,500 images captured from microscopic views of network connectors, which were divided into 70% for training, 15% for validation, and 15% for testing. The proposed model, leveraging a single-stage architecture and attention mechanisms during learning, achieved highly accurate wire detection with approximately 98% precision. Additionally, the overall mean accuracy, classification precision, and recall were around 95%, 99%, and 98%, respectively. The results demonstrate that this system can reliably and in real time verify the correctness of wire color sequencing on the production line without the need for human intervention, thereby reducing human error and enhancing efficiency in the manufacturing process.

**Keywords:** automated inspection, wire color-sequence detection, deep learning, computer vision, YOLO, network cable, production quality control

## 1. Introduction

In the industrial production process of network cables, the correct arrangement of wire strands inside the connector is of critical importance, as any transposition in the color sequence of these wires will result in the production of defective and non-functional cables. Under wiring standards such as T568A and T568B, the sequence of the eight wire strands connected to the network through the connector is precisely defined, and any error at this stage may lead to complete disruption of network performance (Tian et al., 2025). In many production lines, however, verification of the correctness of these color sequences is still performed manually by human operators, who visually assess wire colors using optical magnification (often through digital microscopes). This method is not only time-consuming, but is also subject to operator fatigue, ambient lighting, individual skill, and limitations of human visual acuity, and is therefore prone to error (Wang et al., 2024).

To reduce human errors, increase speed, and achieve uniform quality control, the use of automatic optical inspection[1] systems has been developed across many industrial production domains. The need to employ intelligent and control systems instead of manual supervision is not limited to the cable manufacturing industry, but has also been used in other sensitive applications such as medical equipment for injury prevention and precise pressure management through similar approaches, which further highlights the importance of developing such systems (Shayan et al., 2021). In this context, combining machine vision[2] with deep learning[3] has emerged as an effective and reliable solution across various domains (Yuan et al., 2025; Zhao et al., 2024). In the domain of detecting and evaluating the sequence of color-coded wires, the use of convolutional neural networks[4] enables precise analysis of images and extraction of color and positional features of wires (Liu et al., 2025). Although the development of these systems requires an adequate volume of training data and high processing power for model training, when properly designed they can be deployed in real production lines in real time[5] (Wang et al., 2024).

Among deep learning algorithms, the YOLO family of models is considered one of the most widely used and fastest methods for instantaneous object detection in images (Redmon et al., 2016). YOLO models, owing to their single stage architecture, provide a highly desirable trade-off between speed and accuracy compared to more accurate but slower methods. This characteristic has made YOLO an

[1] automatic optical inspection
[2] machine vision
[3] deep learning
[4] convolutional neural networks
[5] real-time

attractive option for industrial applications including automated inspection on production lines (Li et al., 2025). In the latest versions of this model, particularly the twelfth version (YOLOv12), through the use of attention mechanisms,[6]multi-scale feature aggregation modules, and loss function[7] optimization, detection accuracy has been significantly improved even for small or visually similar objects (Tian et al., 2025). This version of YOLO, while maintaining very high processing speed (in the range of a few milliseconds), is capable of delivering accuracy at the level of more complex and heavyweight models; it is therefore a suitable option for high-volume industrial applications (Mao & Hong, 2025; Nguyen et al., 2024).

Given the importance of correctly identifying wire color and sequence in the final quality of network cables, and the limitations of traditional visual inspection methods, this study designed and implemented a machine vision system based on YOLOv12. Additionally, multimodal learning[8]approaches that combine visual data with textual data or other sensors are also becoming a standard for improving accuracy in sensitive domains such as medical diagnosis (Alaei et al., 2025). The objective of this study is to develop an accurate, fast, and automated algorithm for evaluating the color arrangement of wires in RJ45 connectors by training the model on real data generated on the cable manufacturing line. After identifying the wires in an image, the system compares their sequence from left to right and determines whether it conforms to the required standard pattern. The following sections present the related work review, methodology, data preparation, model structure, evaluation results, and performance analysis.

## 2. Review of Prior Work

The automatic detection of wire color sequences in multistrand cables is one of the precise technical challenges in industrial quality control that has been less studied as an independent topic and has often been addressed as a component of cable or wiring inspection systems. One of the earliest attempts in this area was made by Tsai and Cheng [9] (2022). They presented an automatic optical inspection system based on classical image processing for detecting the color sequence of wire strands, which extracted wire position and color sequence by analyzing horizontal edges and gradients in images and then applying sequential decision logic. Although this method achieved an accuracy of approximately 98.5% under controlled conditions, it was highly dependent on lighting quality and image sharpness; as the visible wire length decreased or the background color changed, accuracy dropped to below 93%. Moreover, its low processing speed (approximately 1.7 seconds per image) made it impractical for fast production lines (Tsai & Cheng, 2022).

In contrast, methods based on deep learning and in particular convolutional neural networks have been able to over-
come many of the limitations of traditional approaches. Furthermore, the application of deep neural networks to the recognition of spatial patterns and position estimation, even in three-dimensional space and for complex gestures, demonstrates the high capability of these models in extracting precise spatial information (Mirzadeh & Zare, 2023). These models have greater generalization power across diverse conditions through learning of spatial and color features of wires. For example, Zhao et al.[10] (2024) used a combination of convolutional networks and attention layers to detect wire sequences in power cables, achieving an accuracy of over 96% with real-time processing speed. In another study, Yuan et al.[11] (2025) presented a system based on YOLOv8 designed for detecting wire connections to electrical terminals, which was able to detect the presence or absence of a wire in the target position with an accuracy exceeding 99%.

In addition to similar applications in wiring harnesses, YOLO models with various modifications have also been widely used in the inspection of wire and industrial cable surfaces. In the work of Mao and Hong [12] (2025), the application of YOLO for detecting defects in fabric textures and industrial wires was studied. Using various YOLO versions (from version 1 to 11), they concluded that detection accuracy for defects continuously improved with each version upgrade and the addition of attention mechanisms.

From a technical standpoint, the evolution of the YOLO model from version 1 to 12 demonstrates remarkable growth in its capabilities for real-time and precise applications. Version 2 introduced anchor boxes [13] with automatic dimension determination, improving localization accuracy over version 1 (Redmon & Farhadi, 2017). Versions 3 and 4, using multi-scale network architectures such as PANet and innovative data augmentation techniques such as Mosaic, achieved higher accuracy in detecting small and visually similar objects (Bochkovskiy et al., 2020). Subsequently, versions 5 through 8 (YOLOv5 to YOLOv8) focused on optimizing speed, reducing model size, and eliminating anchor dependency, developing anchor-free[15] structures that were both more flexible and simpler to train (Jocher et al., 2023).

In the most recent developments, Tian et al.[16] (2025) presented YOLOv12 with a focus on integrating convolutional vision and attention mechanisms. By adding regional attention (A2) modules, aggregation structures such as ELANResidual, and memory optimizations such as

[6] attention mechanisms
[7] loss function
[8] multimodal learning
[9] Tsai & Cheng
[10] Zhao et al.
[11] Yuan et al.
[12] Mao & Hong
[13] anchor boxes
[15]anchor-free
[16]Tian et al.

FlashAttention, this model provides both higher accuracy than previous versions and very low inference time (in the range of 1–2 milliseconds). For example, the YOLOv12-Nano version on the standard COCO dataset has successfully achieved a mean average precision exceeding 40% (Nguyen et al., 2024).

The research of Li et al.[14] (2025) demonstrated that YOLOv12 can also be deployed at industrial scale, offering 97.8% accuracy in detecting complex color patterns and processing speeds exceeding 120 frames per second in field trials. Consequently, combining this version of YOLO with real data can provide a suitable solution for the automated inspection of wire color sequences in network cable production lines.

## 3. Methodology

In this section, the study's methodology—dataset, selected model, and chosen working environment—will be examined.

### 3.1. Training Data and Preprocessing

To train and evaluate the proposed model, a dedicated dataset of images of standard network cable heads was prepared. For this purpose, 2,500 color images with a resolution of approximately 1200×1600 pixels were captured using an industrial digital microscope from a front view of network connectors (RJ45). Each image contained a clear view of eight color-coded wire strands positioned inside the transparent RJ45 socket, which, according to the T568B standard, comprised the colors white-orange, orange, white-green, blue, white-blue, green, white-brown, and brown.

In the next stage, all images were manually annotated. A bounding box[15] was drawn around each wire strand and the corresponding color label was recorded. In total, eight color classes were defined, and the data were split at a ratio of 70% for training, 15% for validation, and 15% for testing.

To improve model generalizability and prevent overfitting, [16] data augmentation [17] techniques were employed. These operations included slight rotation, limited cropping and zooming, horizontal flipping, brightness variation, and contrast adjustment of images. These measures enable the model to adapt well to appearance variations in real production line conditions. Figure 1 shows a sample image captured from the cables.

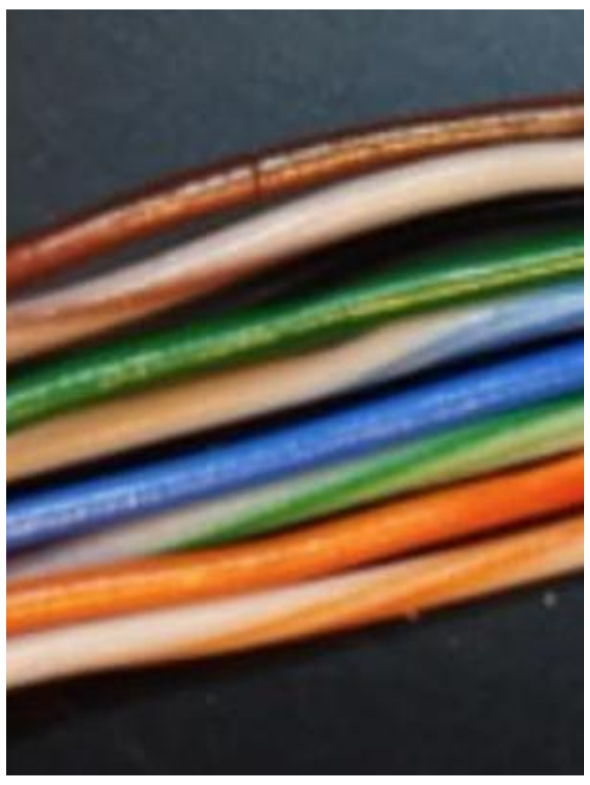

Figure 1: Sample image captured from cables and their arrangement in the laboratory setup

### 3.2. Model Architecture and Training Settings

The model selected for this study was YOLOv12, [18] implemented as an open-source library within the PyTorch [19] framework. YOLOv12 utilizes a single-stage architecture for object detection, directly outputting bounding box coordinates and object classes in a single pass.

The structure of this model comprised three main components:

1. Feature extraction (backbon[20]), in which a regionbased attention[21] mechanism was used to focus on important regions of the image;
2. Feature aggregation using a residual-ELAN[22] structure to preserve information at network depth and improve gradients during training;
3. Multi-scale prediction, in which objects of various sizes—including thin wires—were identified at three output levels of the network.

To adapt the model to the intended application, the model output was configured to recognize eight wire color classes. Although wire sizes in the images were approximately uniform, activating three output scales improved accuracy under varying conditions. Additionally, anchor boxes[23] were used, optimized via k-means clustering[24] based on the actual dimensions of the wires. Ultimately, nine anchor boxes were distributed across three network levels to achieve more precise localization.

The model's loss function comprised three components:

**1. Localization loss** (using the Complete Intersection over Union, CIoU, function) to minimize error in predicted coordinates:

[14] Li et al.
[15] bounding box
[16] overfitting
[17] data augmentation
[18] YOLOv12
[19] pytorch
[20] backbone
[21] region-based attention
[22] residual-ELAN
[23] anchor boxes
[24] k-means clustering

$$L_{CIoU} = 1 - IoU + \frac{\rho^2(b, b^{gt})}{c^2} + \alpha v \quad (1)$$

In this study, the CIoU loss function was used to reduce localization error. As shown in Eq. (1), in addition to the overlap area ($IoU$) between the predicted box ($b$) and the ground-truth box ($b^{gt}$), it also accounts for the distance between their centers ($\frac{\rho^2(b, b^{gt})}{c^2}$) and the stability of the aspect ratio (the term $\alpha v$). This approach leads to faster and more accurate convergence in coordinate prediction compared to more traditional methods.

**2. Objectness loss** based on binary cross-entropy to determine the presence or absence of an object at each candidate position.

**3. Classification loss** for determining the correct wire color, which was enhanced with a label smoothing technique. Successful classification requires a precise correspondence between the visual features extracted by the model and the semantic labels (color); a topic that has also been investigated in recent research on semantic image representation (Zare, 2025).

The weighted combination of these components was set so that approximately 50% was allocated to localization, 40% to objectness detection, and 10% to classification:

$$L_{Total} = \lambda_{Loc} L_{CIoU} + \lambda_{Obj} L_{Objectness} + \lambda_{Cls} L_{Classification} \quad (2)$$

The final loss function $L_{Total}$, minimized during the optimization process, is a weighted combination of the three main components shown in Eq. (2). The weight coefficients ($\lambda$) determine the importance of each component. Accordingly, the model's primary focus was placed on localization accuracy with $\lambda_{Loc} = 0.5$, followed by object detection capability with $\lambda_{Obj} = 0.4$, and finally classification accuracy of wire colors with $\lambda_{Cls} = 0.1$. This weight distribution reflects the study's prioritization of localization and detection precision.

The main evaluation metric during training was the mean average precision [25] on the validation set at various overlap[26]thresholds:

$$mAP = \frac{1}{C}\sum_{i=1}^{C} AP_i \quad (3)$$

The main performance evaluation metric was the mean average precision (mAP) at a specified overlap threshold ($mAP@IoU$), defined in Eq. (3). $C$ is the total number of color classes (equal to 8) and $AP_i$ represents the Average Precision for class $i$, which is the area under the precisionrecall curve for that class. Using mAP ensures that the model achieves good performance in both correctly identifying (high precision) and finding all relevant instances (high recall) for all eight wire colors.

### *3.3. Training Process*

Model training was performed in Google Colab[27] using an NVIDIA Tesla T4[28] GPU. Initial weights were loaded from a YOLO model pre-trained on the COCO[29] public dataset so that training would begin from an optimal starting point.

The initial learning rate was set to 0.001, which was reduced in a step-wise manner over the course of 50 training epochs.[30] The batch size[31] was set to 16 due to graphics memory constraints. At the end of each epoch, the model's performance on the validation data was recorded, and learning curves showing the trend of loss function decrease and accuracy increase were plotted. Furthermore, to prevent overfitting, if mAP did not improve over five consecutive epochs, training was automatically stopped (early stopping[32]). Training time per epoch was estimated at approximately 1 to 2 minutes, with total training time estimated at approximately one hour.

### *3.4. Decision Algorithm and Standard Pattern Matching*

After training completion, a simple but precise algorithm was designed for practical use on the production line to evaluate wire color sequences. In this algorithm, the model output—including the spatial coordinates and color of each wire strand—is first extracted. The color sequence is then obtained by sorting the wires based on the horizontal component of their position (x-axis).

In the next step, the predicted sequence is compared against the required standard pattern (e.g., T568B). If a complete match exists, the cable is considered healthy and correctly wired; otherwise, the cable is labeled as defective and set aside for further inspection. Due to its simplicity and high speed, this process adds no significant computational overhead relative to the detection stage and is capable of real-time execution on the production line.

### *3.5. Implementation Environment and Tools*

All training and evaluation stages were conducted in the Google Colab environment. Hardware specifications included an NVIDIA Tesla T4 GPU with 16 GB of memory, a 4-core virtual CPU, and 12 GB of runtime memory. For model implementation, the open-source Ultralytics[33] library for YOLOv12 was used, which provides model loading,

[25] Mean average precision (mAP)
[26] IoU
[27] google colab
[28] nvidia tesla T4
[29] COCO
[30] epoch
[31] batch size
[32] early stopping
[33] ultralytics

training, parameter tuning, and evaluation in a simple framework.

For image processing and data augmentation, the OpenCV[34] and Albumentations[35] libraries were also used. For plotting and analyzing learning trends, the Matplotlib[36] library was employed. The implementation code was designed in a modular fashion to facilitate easy deployment in industrial environments. The model's processing speed after training was estimated at an average of 5–6 milliseconds per image, equivalent to more than 160 frames per second. This speed is several times faster than the typical requirements of industrial production lines. Even running on a CPU (without a GPU accelerator), the model was executable at approximately 8–9 frames per second, which is also suitable for offline and testing applications.

## 4. Results and Evaluation

After training the YOLOv12 model and optimizing its parameters, the model's performance was evaluated on a set of 375 test images that were completely independent of the training and validation data. These images included samples from network cables under varied conditions (such as lighting variation, noise presence, and background differences) to assess the model's generalization ability on unseen data.

Figure 2 shows the trend of loss function decrease and mAP accuracy increase of the model on the validation data over 50 training epochs. As can be observed, the model progressed uniformly toward convergence and achieved an accuracy of $mAP \approx 95\%$ after approximately 40 epochs, indicating stable training without obvious signs of overfitting. Data augmentation techniques, early stopping, and the relative simplicity of color classification were among the factors contributing to training stability.

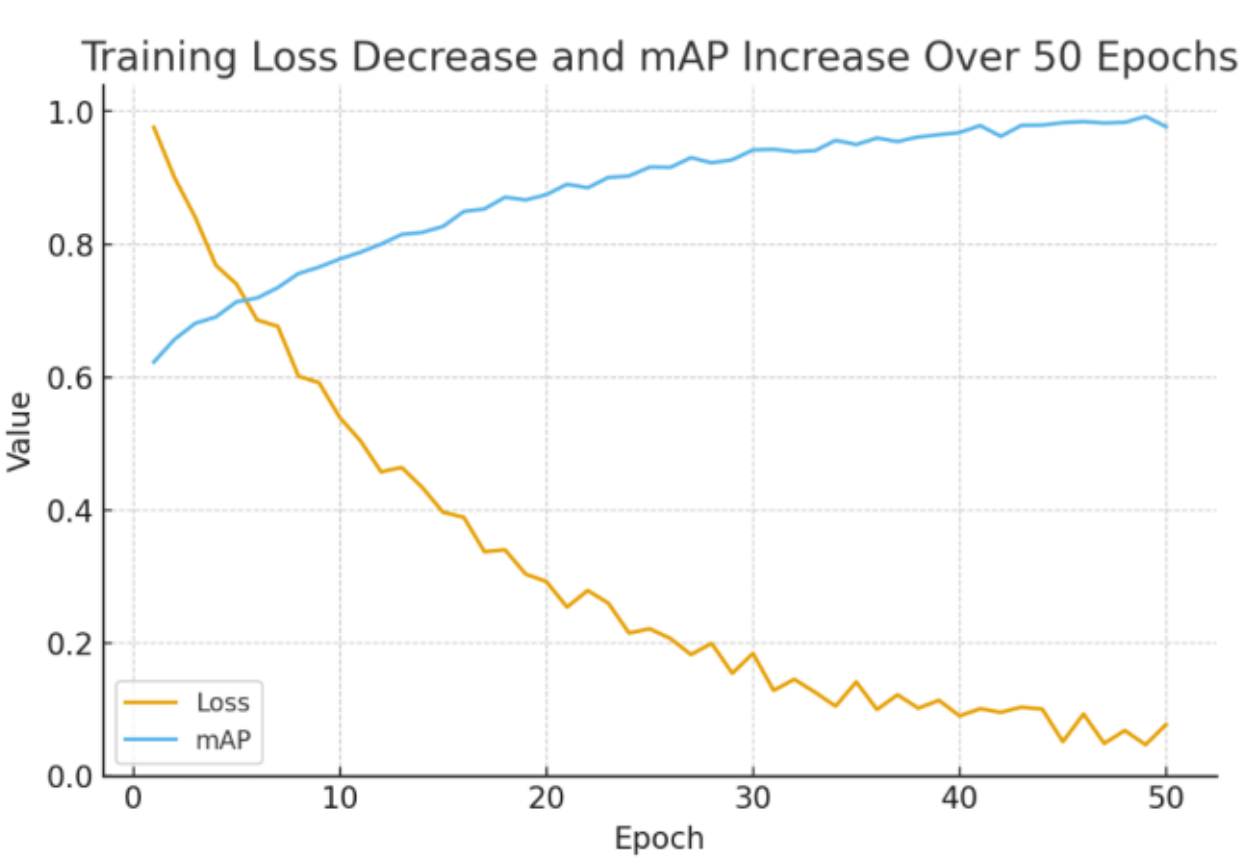

Figure 2: Trend of YOLO model loss and mAP accuracy during the training process

In the final evaluation, the model was able to detect fewer than 7 out of a total of 8 wires in the test images on only rare occasions. In most cases, all 8 wires were correctly identified and received the appropriate color label. The quantitative evaluation metrics are as follows:

- Precision: 99.1%
- Recall: 98.0%
- mAP (IoU = 0.5:0.95): 94.6%

Figure 3 shows the model's confusion matrix, which indicates that the most common errors were related to wires with white insulation and narrow color stripes (such as whiteorange and white-brown). These errors, totaling less than 2%, were caused by color similarity and low-contrast differences in some images. Solid-color wires such as orange, blue, green, and brown were classified with accuracy close to 100%.

Qualitative review of the model output also shows that the left-to-right wire sequence in virtually all samples was consistent with the standard pattern. Even in the few cases where the color label was incorrect, the physical position of the wire was at the correct location in the sequence. Consequently, in the sequence verification check, the model was able to correctly declare 368 out of 375 cables as pass/fail (final accuracy: 98.13%). It is worth noting that only 7 false reject[37] errors were observed, while no false accept[38] errors were detected.

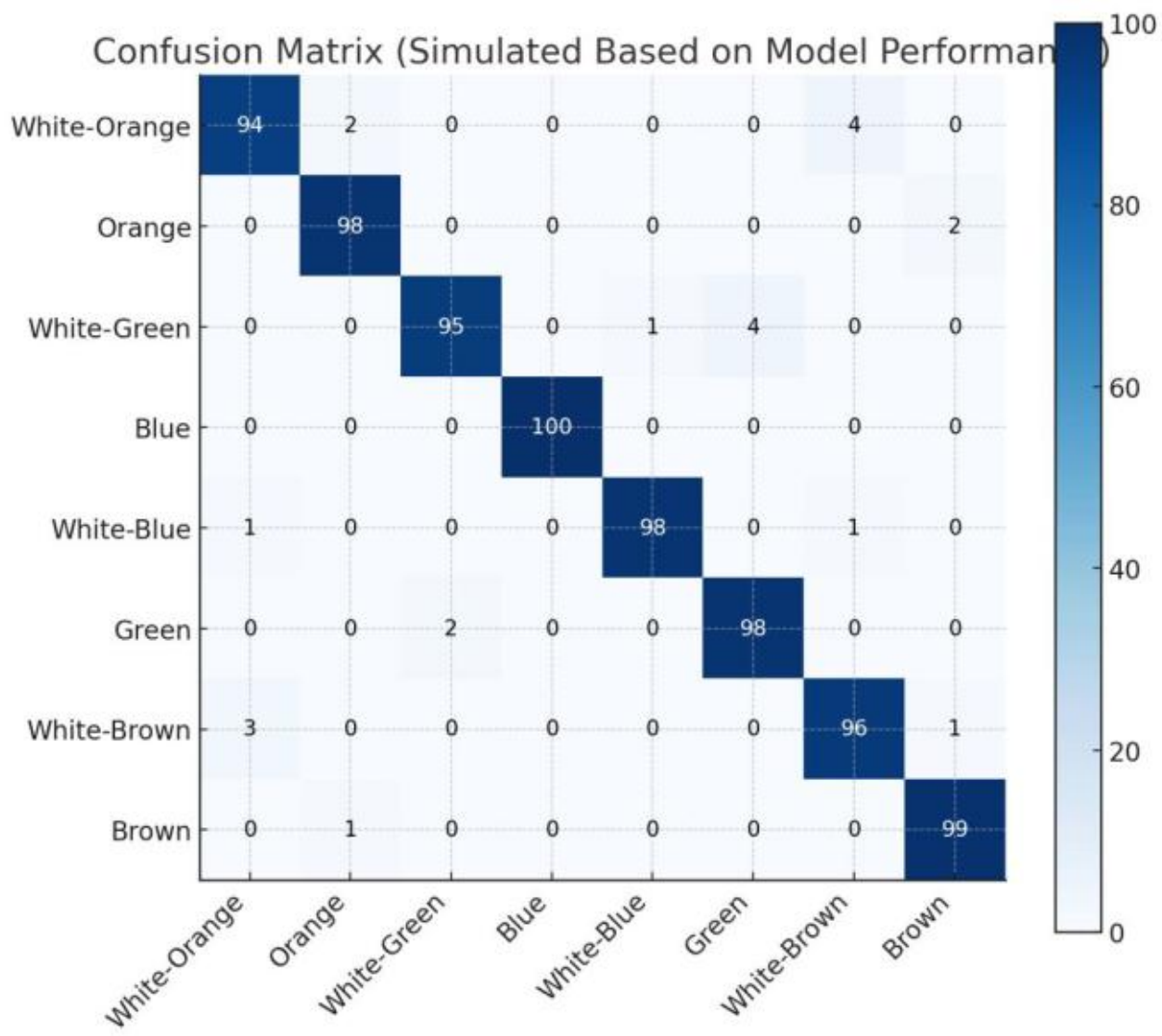

### 4.1. Comparison with Prior Methods

The results show that YOLOv12 has a notable superiority over traditional image processing methods in terms of accuracy, stability, and speed. For example, the method of Tsai and Cheng (2022) achieved an accuracy of 98.5% under controlled conditions, but operated at less than 1 frame per second.[39] In contrast, the present model was able

[34] openCV
[35] albumentations
[36] matplotlib
[37] false reject
[38] false accept
[39] fps

to run with similar accuracy and at a speed of approximately 160 fps, which is entirely suitable for real-time applications on production lines.

Furthermore, experiments with different backgrounds (yellow, black) showed that the model primarily focuses on the wires themselves and is robust to environmental variations. This capability—in contrast to traditional methods that are highly dependent on background and lighting— stems from the model's use of deep learned features.

### 4.2. Model Performance Summary

With its advanced architecture and optimized training, YOLOv12 successfully met the requirements for accuracy, speed, and reliability in the network cable inspection application. These results demonstrate the operational viability of its use on industrial production lines. Figure 4 shows a sample output of the model.

## 5. Discussion and Conclusion

In this study, the design and implementation of an automated system for detecting and verifying the color sequence of wires in network cables based on the advanced YOLOv12 architecture was investigated. The results showed that using deep learning for this problem can offer significant advantages in terms of accuracy, speed, and reliability over traditional methods and visual inspection. The developed model was able to correctly evaluate wire strand sequences with an accuracy exceeding 98%—a level of accuracy not only entirely suitable for industrial environments, but also surpassing human error.

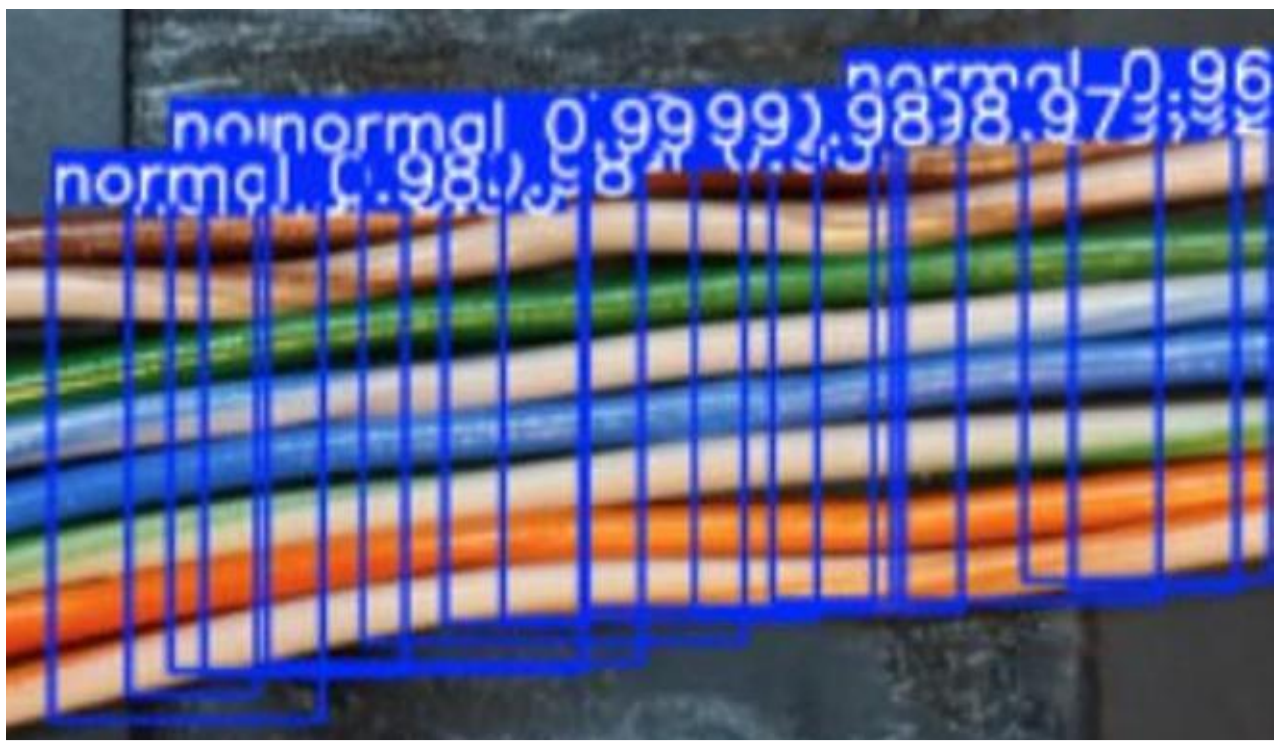

Figure 4: Sample model output showing color identification and verification of correct color sequence

Beyond accuracy, the high speed of the system is also one of its key advantages. Real-time image processing enables cables to be evaluated immediately after production and accepted or rejected without creating a bottleneck. This characteristic is fully aligned with the requirements of Industry 4.0 automation and the goal of 100% quality control in modern production lines. From a technical standpoint, the use of YOLOv12—in which optimization techniques such as regional attention and hierarchical feature fusion are implemented—ensured that precise detection of wires with small dimensions and high color similarity (such as striped colors) was also performed well.

Nevertheless, challenges did exist. Manual annotation of the data and the preparation of 2,500 labeled images was a time-consuming process. In future developments, the use of synthetic data through simulation tools or augmented reality can reduce the cost of data collection.

Complementary approaches such as transfer learning can also be effective. Using weights trained on RJ45 cables, the model can be retrained for similar applications with less data. This would be useful for extending the system to other cable types, such as telephone or industrial cables. From an industrial deployment standpoint, two main options exist: leveraging centralized GPU servers or deploying on embedded hardware such as the Nvidia Jetson. Lightweight versions of YOLOv12 such as Nano, with only 2.6 million parameters, are also capable of running on moderately priced systems. However, practical testing on such platforms is necessary to ensure stable performance in an industrial environment.

For further improving final accuracy, multi-stage systems can be employed. For example, after initial detection by YOLO, a more precise network such as a Vision Transformer could be applied to each region to detect wire color with higher accuracy. Moreover, combining visual processing with electronic measurements such as wire continuity testing can help increase confidence, particularly in cases where visual data is ambiguous.

Overall, the results of this study have shown that the YOLOv12-based system acts not only as an effective replacement for manual network wire inspection in terms of accuracy, but also in terms of speed, stability, and flexibility. Deploying such a system on production lines, while reducing waste from defective cables, will lead to reduced manpower requirements and improved product quality uniformity. Furthermore, this approach is extensible to other applications such as inspection of automotive wire harnesses, electronic board connections, and other multi-strand systems. Finally, by connecting this system to robotic production lines, an important step toward the implementation of smart factories will have been taken.

---

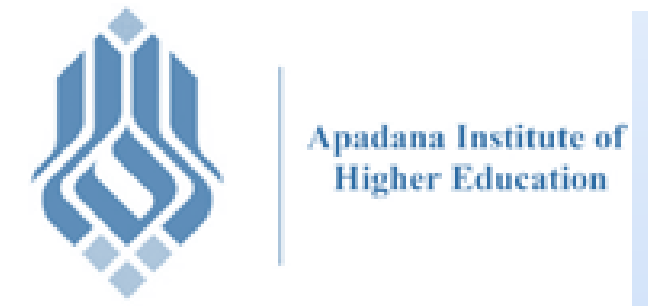

# Using the YOLOv12 Model for Verifying the Correct Color Sequence of Wires in Network Cables (Patch Cords) on the Production Line

**Amin Doroodchi*** | Computer Department, Islamic Azad University, Beyza branch, Beyza, Iran

**Danial Soleymani** | R&D at Nedaye Sabz Company, Isfahan Branch, Iran

## Abstract

In the production process of network cables, ensuring the correct color sequence of wire pairs inside the standard connector plays a critical role in the final performance of the cable, as any misplacement or color-ordering error can lead to defective products and impose significant costs. Traditional inspection methods based on visual examination through digital microscopes are typically time-consuming, tedious, and prone to human error. In this study, an intelligent system based on the twelfth version of the YOLO object detection model was developed to identify the position and verify the correct color sequence of wires in patch cords. The dataset used consisted of 2,500 images captured from microscopic views of network connectors, which were divided into 70% for training, 15% for validation, and 15% for testing. The proposed model, leveraging a single-stage architecture and attention mechanisms during learning, achieved highly accurate wire detection with approximately 98% precision. Additionally, the overall mean accuracy, classification precision, and recall were around 95%, 99%, and 98%, respectively. The results demonstrate that this system can reliably and in real time verify the correctness of wire color sequencing on the production line without the need for human intervention, thereby reducing human error and enhancing efficiency in the manufacturing process.

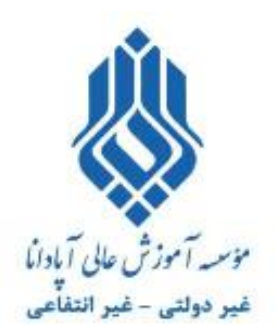

# استفاده از مدل YOLOv12 برای تشخیص صحت ترتیب رنگ سیم‌ها در کابل‌های شبکه (پچ‌کورد) در خط تولید

**امین درودچی*** | گروه کامپیوتر، دانشگاه آزاد اسلامی واحد بیضا، بیضا، ایران

**دانیال سلیمانی** | واحد تحقیق و توسعه، شرکت ندای سبز، اصفهان، ایران

## چکیده

در فرآیند تولید کابل‌های شبکه، اطمینان از صحت ترتیب رنگ رشته‌سیم‌ها در داخل کانکتور استاندارد نقش بسیار مهمی در عملکرد نهایی کابل دارد، چرا که هرگونه جابه‌جایی یا اشتباه در رنگ‌بندی می‌تواند منجر به تولید محصول معیوب و تحمیل هزینه‌های قابل توجه شود. روش‌های سنتی بازرسی که مبتنی بر مشاهده چشمی از طریق میکروسکوپ دیجیتال هستند، معمولاً زمان‌بر، خسته‌کننده و مستعد خطاهای انسانی هستند. در این پژوهش، یک سامانه هوشمند بر پایه مدل شناسایی اشیای نسخه دوازدهم یولو[1] به منظور تشخیص موقعیت و ترتیب رنگ صحیح رشته‌سیم‌ها در کابل‌های پچ‌کورد توسعه داده شده است. مجموعه داده مورد استفاده شامل دو هزار و پانصد تصویر تهیه‌شده از نمای میکروسکوپی کانکتورهای شبکه بود که به نسبت هفتاد درصد برای آموزش، پانزده درصد برای اعتبارسنجی و پانزده درصد برای آزمون تقسیم شدند. مدل پیشنهادی با بهره‌گیری از ساختار تک‌مرحله‌ای و مکانیزم توجه در مرحله یادگیری، موفق به شناسایی دقیق سیم‌ها با دقت حدود نود و هشت درصد گردید. همچنین میانگین دقت کلی ، دقت طبقه‌بندی و نرخ بازیابی به ترتیب در حدود نود و پنج درصد، نود و نه درصد و نود و هشت درصد به دست آمدند. نتایج نشان می‌دهد که این سامانه می‌تواند به صورت بلادرنگ، قابل اعتماد و بدون نیاز به مداخله انسانی، درستی ترتیب رنگ سیم‌ها را در خط تولید بررسی کرده و موجب کاهش خطاهای انسانی و افزایش بهره‌وری در فرآیند تولید گردد.

---
[1] YOLOv12

**مقدمه**

در فرآیند تولید صنعتی کابل‌های شبکه، صحت چیدمان رشته‌سیم‌ها در داخل کانکتور از اهمیت حیاتی برخوردار است، چرا که هرگونه جابه‌جایی در ترتیب رنگ این سیم‌ها منجر به تولید کابل‌های معیوب و ناکارآمد خواهد شد. در استانداردهای سیم‌بندی نظیر T568A و T568B، ترتیب هشت‌رشته‌سیمی که از طریق کانکتور به شبکه متصل می‌شوند به‌صورت دقیق تعریف شده و هرگونه خطا در این مرحله ممکن است منجر به اختلال کامل در عملکرد شبکه شود(Tian و همکاران، ۲۰۲۵). این در حالی است که در بسیاری از خطوط تولید، بررسی صحت این ترتیب رنگ‌ها همچنان به صورت دستی و توسط نیروی انسانی انجام می‌شود، به‌گونه‌ای که اپراتورها با کمک بزرگ‌نمایی نوری (اغلب از طریق میکروسکوپ دیجیتال)، رنگ رشته‌ها را به‌صورت چشمی ارزیابی می‌کنند. این روش نه‌تنها زمان‌بر، بلکه تحت تأثیر خستگی اپراتور، نور محیط، مهارت فردی و محدودیت دقت بینایی انسان بوده و در نتیجه مستعد بروز خطا است (Wang و همکاران، ۲۰۲۴).

به‌منظور کاهش خطاهای انسانی، افزایش سرعت و دستیابی به کنترل کیفیت یکنواخت، استفاده از سامانه‌های بازرسی خودکار نوری[1] در بسیاری از حوزه‌های تولید صنعتی توسعه یافته است. البته لزوم به‌کارگیری سیستم‌های کنترلی و هوشمند به جای نظارت دستی، تنها محدود به صنعت کابل‌سازی نیست؛ بلکه در سایر کاربردهای حساس مانند تجهیزات پزشکی نیز برای پیشگیری از آسیب و مدیریت دقیق فشار، از رویکردهای مشابهی استفاده شده است که اهمیت توسعه چنین سامانه‌هایی را دوچندان می‌کند (Shayan و همکاران، ۲۰۲۱). در این میان، ترکیب روش‌های بینایی ماشین[2] با یادگیری عمیق[3] به راهکاری مؤثر و قابل اعتماد در حوزه‌های مختلف تبدیل شده است (Yuan و همکاران، ۲۰۲۵؛ Zhao و همکاران، ۲۰۲۴). در حوزه تشخیص و ارزیابی ترتیب سیم‌های رنگی نیز به‌کارگیری شبکه‌های عصبی کانولوشنی[4] امکان تحلیل دقیق تصاویر، استخراج ویژگی‌های رنگ و موقعیت مکانی سیم‌ها را فراهم می‌سازد (Liu و همکاران، ۲۰۲۵). اگرچه توسعه این سامانه‌ها نیازمند حجم مناسبی از داده‌های آموزشی و قدرت پردازشی بالا برای آموزش مدل‌ها است، اما در صورت طراحی صحیح می‌توان آن‌ها را در خطوط تولید واقعی و به‌صورت بلادرنگ[5] پیاده‌سازی کرد (Wang و همکاران، ۲۰۲۴).

در میان الگوریتم‌های یادگیری عمیق، خانواده مدل‌های یولو یکی از پرکاربردترین و سریع‌ترین روش‌ها برای تشخیص آنی اشیاء در تصویر محسوب می‌شود (Redmon و همکاران، ۲۰۱۶). مدل‌های یولو به دلیل معماری تک‌مرحله‌ای خود، در مقایسه با روش‌های دقیق‌تر اما کندتر، توازن بسیار مطلوبی بین سرعت و دقت فراهم می‌آورند. این ویژگی موجب شده است که یولو به گزینه‌ای جذاب برای کاربردهای صنعتی از جمله بازرسی خودکار در خطوط تولید تبدیل شود (Li و همکاران، ۲۰۲۵). در نسخه‌های اخیر این مدل، به‌ویژه نسخه دوازدهم (یولو ۱۲)، با استفاده از مکانیزم‌های توجه[6]، ماژول‌های تجمیع ویژگی‌های چندمقیاسی و بهینه‌سازی توابع اتلاف[7]، دقت تشخیص حتی برای اشیای کوچک یا مشابه به‌طور چشمگیری افزایش یافته است (Tian و همکاران، ۲۰۲۵). این نسخه از یولو ضمن حفظ سرعت پردازش بسیار بالا (در حد چند میلی‌ثانیه)، قادر است دقتی در سطح مدل‌های پیچیده‌تر و سنگین‌تر ارائه دهد،

---
[1] Automatic Optical Inspection
[2] Machine Vision
[3] Deep Learning
[4] Convolutional Neural Networks
[5] Real-Time
[6] Attention Mechanisms
[7] Loss Functions

و از این‌رو گزینه‌ای مناسب برای کاربردهای صنعتی با حجم تولید بالا به‌شمار می‌رود (Mao و Hong، ۲۰۲۵؛ Nguyen و همکاران، ۲۰۲۴).

با توجه به اهمیت تشخیص صحیح رنگ و ترتیب سیم‌ها در کیفیت نهایی کابل شبکه، و محدودیت‌های روش‌های سنتی بازرسی چشمی، در این پژوهش یک سامانه بینایی ماشین مبتنی بر مدل یولو نسخه دوازدهم طراحی و پیاده‌سازی شده است. همچنین رویکردهای یادگیری چندحالتی[1] که داده‌های دیداری را با داده‌های متنی یا دیگر حسگرها ترکیب می‌کنند، نیز در حال تبدیل شدن به یک استاندارد برای افزایش دقت در حوزه‌های حساسی مانند تشخیص پزشکی هستند (Alaei و همکاران، ۲۰۲۵). هدف از این پژوهش آن است که بتوان از طریق آموزش مدل بر پایه داده‌های واقعی تولیدشده در خط کابل‌سازی، یک الگوریتم دقیق، سریع و خودکار برای ارزیابی چیدمان رنگ سیم‌ها در کانکتورهای RJ45 توسعه داد. این سامانه پس از شناسایی سیم‌ها در تصویر، ترتیب آن‌ها را از چپ به راست مقایسه کرده و صحت انطباق آن با الگوی استاندارد را تعیین می‌کند. در ادامه، ضمن مرور پژوهش‌های مرتبط، روش کار، نحوه آماده‌سازی داده‌ها، ساختار مدل، نتایج ارزیابی و تحلیل عملکرد ارائه خواهد شد.

## مرور پژوهش‌های پیشین

تشخیص خودکار ترتیب رنگ سیم‌ها در کابل‌های چندرشته‌ای یکی از چالش‌های فنی دقیق در حوزه کنترل کیفیت صنعتی است که به‌صورت تخصصی کمتر در مطالعات مستقل بررسی شده و اغلب در قالب بخشی از سامانه‌های بازرسی کابل یا سیم‌بندها مورد توجه قرار گرفته است. یکی از نخستین تلاش‌ها در این زمینه را تسای و چنگ (Tsai & Cheng, 2022) انجام دادند. آن‌ها یک سامانه بازرسی خودکار نوری مبتنی بر پردازش تصویر کلاسیک برای تشخیص توالی رنگ رشته‌سیم‌ها ارائه کردند که با تحلیل لبه‌ها و گرادیان‌های افقی در تصاویر و سپس استفاده از منطق تصمیم‌گیری ترتیبی، توانستند محل و ترتیب رنگ سیم‌ها را استخراج کنند. این روش اگرچه در شرایط کنترل‌شده دقتی در حدود ۹۸٫۵ درصد داشت، اما به شدت به کیفیت نور و وضوح تصویر وابسته بود، به‌طوری‌که با کاهش طول نمای قابل رؤیت سیم‌ها یا تغییر رنگ پس‌زمینه، دقت تا کمتر از ۹۳ درصد افت می‌کرد. همچنین سرعت پردازش پایین (حدود ۱٫۷ ثانیه برای هر تصویر) آن را برای خطوط تولید سریع غیرکاربردی می‌کرد (Tsai & Cheng, 2022).

در مقابل، روش‌های مبتنی بر یادگیری عمیق و به‌ویژه شبکه‌های عصبی کانولوشنی توانسته‌اند بر بسیاری از محدودیت‌های روش‌های سنتی غلبه کنند. همچنین، کاربرد شبکه‌های عصبی عمیق در تشخیص الگوهای مکانی و تخمین موقعیت، حتی در فضای سه‌بعدی و برای ژست‌های پیچیده، نشان‌دهنده توانایی بالای این مدل‌ها در استخراج اطلاعات دقیق مکانی است (Mirzadeh و Zare، ۲۰۲۳). این مدل‌ها با یادگیری ویژگی‌های مکانی و رنگی سیم‌ها، قدرت تعمیم بالاتری به شرایط متنوع دارند. برای مثال، ژائو و همکاران (Zhao et al., 2024) از ترکیب شبکه‌های کانولوشنی و لایه‌های توجه برای تشخیص ترتیب سیم‌ها در کابل‌های برق استفاده کردند و به دقتی فراتر از ۹۶ درصد با سرعت پردازش بلادرنگ دست یافتند. در پژوهشی دیگر، یوان و همکاران (Yuan et al., 2025) سامانه‌ای مبتنی بر یولو نسخه هشتم ارائه کردند که برای تشخیص اتصال سیم به ترمینال‌های الکتریکی طراحی شده بود و توانست وجود یا عدم وجود سیم در موقعیت موردنظر را با دقت بالای ۹۹ درصد تشخیص دهد.

علاوه بر کاربردهای مشابه در سیم‌بندها، در بازرسی سطوح سیم‌ها و کابل‌های صنعتی نیز مدل‌های یولو با اصلاحات مختلف به‌طور گسترده استفاده شده‌اند. در پژوهش مائو و هونگ (Mao & Hong, 2025) کاربرد مدل یولو در

[1] Multimodal Learning

شناسایی عیوب بافت پارچه و سیم‌های صنعتی بررسی شد. آن‌ها با استفاده از نسخه‌های مختلف یولو (از نسخه اول تا یازدهم) به این نتیجه رسیدند که دقت تشخیص نواقص، با ارتقاء نسخه مدل و افزودن مکانیزم‌های توجه، به‌طور پیوسته بهبود یافته است.

از منظر فنی، تکامل مدل یولو از نسخه اول تا دوازدهم نشان‌دهنده رشد قابل‌توجهی در توانمندی‌های آن برای کاربردهای بلادرنگ و دقیق است. نسخه دوم یولو با معرفی جعبه‌های لنگر[1] و تعیین خودکار ابعاد آن‌ها، دقت مکان‌یابی را نسبت به نسخه اول بهبود داد (Redmon & Farhadi, 2017). نسخه‌های سوم و چهارم با بهره‌گیری از معماری شبکه‌های چندمقیاسی مانند PANet و تکنیک‌های داده‌افزایی نوآورانه نظیر Mosaic، به دقتی بالاتر در شناسایی اشیای کوچک و مشابه دست یافتند (Bochkovskiy et al., 2020). در ادامه، نسخه‌های پنجم تا هشتم (YOLOv5 تا YOLOv8) بر بهینه‌سازی سرعت، کاهش حجم مدل و حذف وابستگی به جعبه‌های لنگر متمرکز شدند و ساختارهایی بدون لنگر (Anchor-Free) را توسعه دادند که هم انعطاف‌پذیرتر و هم ساده‌تر برای آموزش بودند (Jocher et al., 2023).

در تازه‌ترین تحولات، تیان و همکاران (Tian et al., 2025) مدل یولو نسخه دوازدهم را با تمرکز بر تلفیق بینایی کانولوشنی و مکانیزم‌های توجه ارائه کردند. این مدل با افزودن ماژول‌های توجه ناحیه‌ای (A2)، ساختارهای تجمیعی نظیر Residual-ELAN و بهینه‌سازی‌های حافظه مانند FlashAttention، هم دقت بالاتری نسبت به نسخه‌های پیشین و هم زمان استنتاج بسیار پایین (در حد ۱ تا ۲ میلی‌ثانیه) فراهم می‌سازد. به عنوان نمونه، نسخه YOLOv12-Nano بر روی مجموعه داده استاندارد COCO موفق به دستیابی به دقت میانگین بالاتر از ۴۰ درصد شده است (Nguyen et al., 2024).

پژوهش لی و همکاران (Li et al., 2025) نشان داد که نسخه دوازدهم یولو حتی در مقیاس صنعتی نیز توانایی پیاده‌سازی دارد و در آزمایش‌های میدانی، دقت ۹۷/۸ درصدی در تشخیص الگوهای رنگی پیچیده و سرعت پردازش بالاتر از ۱۲۰ فریم بر ثانیه را ارائه داده است. در نتیجه، ترکیب این نسخه از یولو با داده‌های واقعی می‌تواند راهکاری مناسب برای بازرسی خودکار ترتیب رنگ سیم‌ها در خط تولید کابل‌های شبکه باشد.

## روش کار

در این بخش روش کار این پژوهش، دیتاست و مدل انتخاب شده به همراه محیط کاری انتخاب شده بررسی خواهد شد.

## داده‌های آموزشی و پیش‌پردازش

برای آموزش و ارزیابی مدل پیشنهادی، یک مجموعه‌داده اختصاصی از تصاویر سر کابل‌های شبکه استاندارد تهیه گردید. بدین منظور تعداد دو هزار و پانصد تصویر رنگی با وضوح تقریبی ۱۲۰۰ در ۱۶۰۰ پیکسل با استفاده از میکروسکوپ دیجیتال صنعتی از نمای روبروی کانکتورهای شبکه (RJ45) ثبت شد. هر تصویر شامل نمایی واضح از هشت رشته‌سیم رنگی قرارگرفته درون سوکت شفاف RJ45 بود که مطابق با استاندارد T568B، رنگ‌های سفید–نارنجی، نارنجی، سفید–سبز، آبی، سفید–آبی، سبز، سفید–قهوه‌ای و قهوه‌ای را شامل می‌شدند.

در مرحله بعد، کلیه تصاویر به‌صورت دستی نشانه‌گذاری شدند. به این شکل که پیرامون هر رشته‌سیم یک جعبه محصورکننده[2] رسم و برچسب مربوط به رنگ آن ثبت شد. در مجموع، هشت کلاس رنگی تعریف گردید و داده‌ها با نسبت هفتاد درصد برای آموزش، پانزده درصد برای اعتبارسنجی و پانزده درصد برای آزمون تقسیم شدند.

[1] Anchor Boxes
[2] Bounding Box

برای افزایش تعمیم‌پذیری مدل و جلوگیری از بیش‌برازش[1]، از روش‌های افزون‌سازی داده[2] بهره گرفته شد. این عملیات شامل چرخش جزئی، برش و بزرگنمایی محدود، معکوس‌سازی افقی، تغییر روشنایی و تنظیم کنتراست تصاویر بود. این اقدامات سبب می‌شود مدل بتواند به‌خوبی با تنوعات ظاهری در شرایط واقعی خط تولید سازگار شود. در تصویر ۱ نمونه ای ا تصویربرداری از کابل ها را مشاهده میکنید.

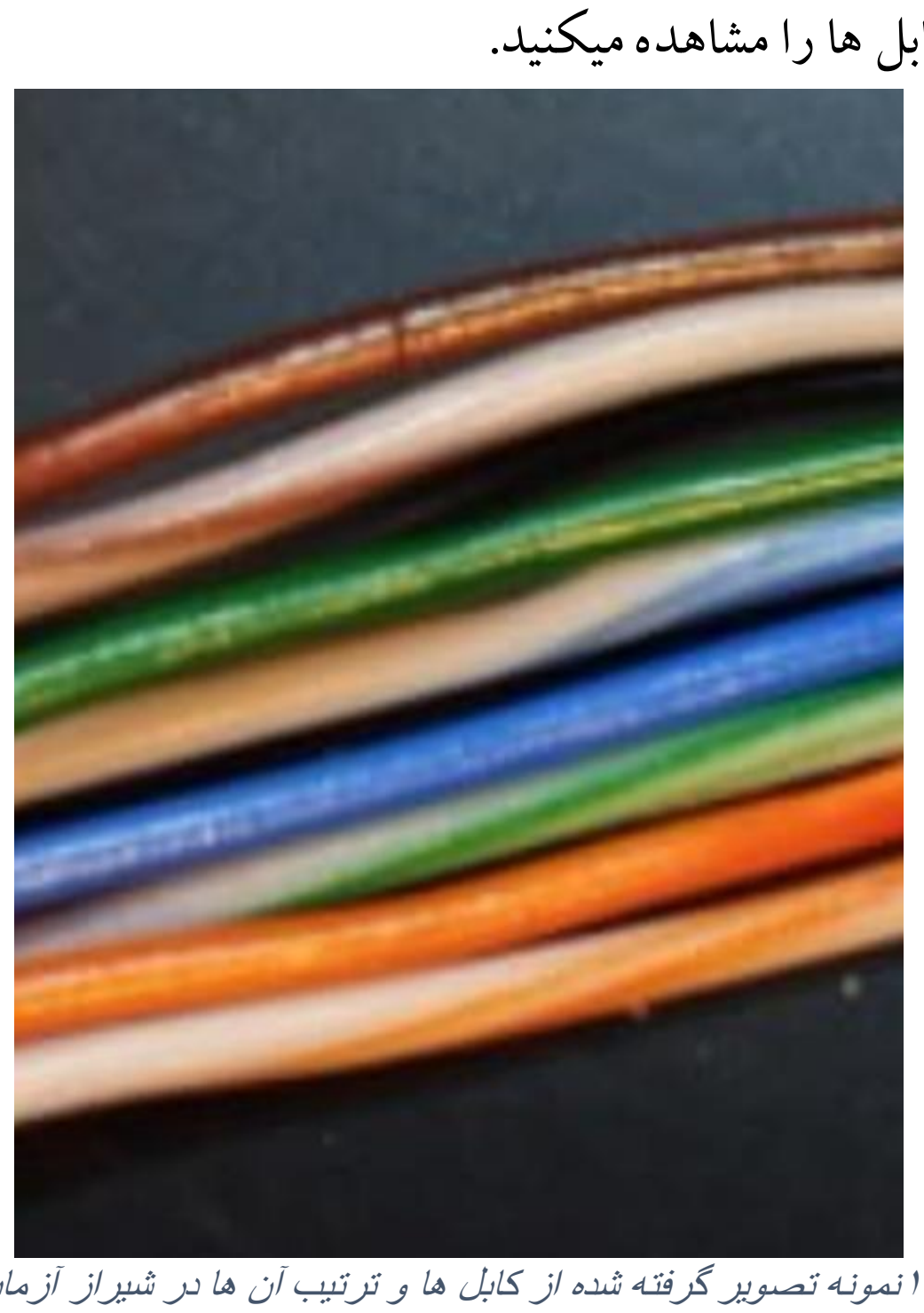

Figure ۱ نمونه تصویر گرفته شده از کابل ها و ترتیب آن ها در شیراز آزمایشگاهی

## معماری مدل و تنظیمات آموزش

مدل انتخاب‌شده‌برای این پژوهش، نسخه دوازدهم مدل یولو۱۲ بود که به‌صورت متن‌باز در بستر پای‌تورچ[3] پیاده‌سازی شده است. یولو نسخه دوازده از معماری تک‌مرحله‌ای برای تشخیص اشیاء بهره می‌برد و مستقیماً مختصات جعبه‌های محصورکننده و کلاس‌های مربوط به اشیاء را در یک مرحله خروجی می‌دهد.

ساختار مدل شامل سه بخش اصلی بود:

1. استخراج ویژگی‌ها[4] که در آن از مکانیزم توجه ناحیه‌ای (Region-based Attention) برای تمرکز بر بخش‌های مهم تصویر استفاده شد؛
2. تجمیع ویژگی‌ها با استفاده از ساختار باقیمانده (Residual-ELAN) ELAN جهت حفظ اطلاعات در عمق شبکه و بهبود گرادیان در فرآیند آموزش؛
3. پیش‌بینی چندمقیاسی که در سه سطح از خروجی شبکه، اشیاء با اندازه‌های مختلف از جمله سیم‌های نازک شناسایی می‌شدند.

برای تطبیق مدل با کاربرد موردنظر، خروجی مدل به گونه‌ای پیکربندی شد که هشت کلاس رنگی سیم را شناسایی کند. اگرچه اندازه سیم‌ها در تصاویر تقریباً یکسان بود، اما فعال‌سازی سه مقیاس خروجی موجب بهبود دقت در شرایط

[1] Overfitting
[2] Data Augmentation
[3] PyTorch
[4] Backbone

مختلف شد. همچنین از جعبه‌های لنگر[1] استفاده گردید که با الگوریتم خوشه‌بندی k-میانگین[2] بر اساس ابعاد واقعی سیم‌ها بهینه‌سازی شدند. در نهایت، نه جعبه لنگر در سه سطح شبکه توزیع گردید تا موقعیت‌یابی دقیق‌تری حاصل شود.

تابع اتلاف مدل شامل سه مؤلفه بود:

1. اتلاف مکان‌یابی[3] با استفاده از تابع هم‌پوشانی کامل CIoU برای کاهش خطا در مختصات پیش‌بینی‌شده؛

$$\boldsymbol{L_{CIoU} = 1 - IoU + \frac{\rho^2(b, b^{gt})}{c^2} + \alpha v}\,(\mathbf{1})$$

در این پژوهش، از تابع اتلاف (Complete Intersection over Union) CIoU برای کاهش خطای مکان‌یابی استفاده شد. این تابع که در فرمول (۱) نمایش داده شده است، علاوه بر اندازه هم‌پوشانی ($IoU$) بین جعبه پیش‌بینی‌شده ($b$) و جعبه حقیقت زمین ($b^{gt}$)، به فاصله مراکز آن‌ها ($\frac{\rho^2(b,b^{gt})}{c^2}$) و همچنین به میزان ثبات در نسبت ابعادی (جمله $\alpha v$) توجه می‌کند. این رویکرد، در مقایسه با روش‌های سنتی‌تر، منجر به همگرایی سریع‌تر و دقیق‌تر در پیش‌بینی مختصات می‌شود.

2. اتلاف تشخیص شیء[4] بر اساس باینری کراس‌انتروپی برای تعیین وجود یا عدم وجود شیء در هر موقعیت پیشنهادی؛

3. اتلاف طبقه‌بندی[5] برای تعیین رنگ صحیح، که با تکنیک نرم‌سازی برچسب[6] بهبود یافته بود. موفقیت در طبقه‌بندی نیازمند ارتباط دقیق بین ویژگی‌های بصری استخراج‌شده توسط مدل و برچسب‌های معنایی (رنگ) است؛ موضوعی که در تحقیقات جدید بر روی نمایش معنایی تصاویر نیز مورد بررسی قرار گرفته است (Zare، ۲۰۲۵).

ترکیب وزنی این مؤلفه‌ها به‌گونه‌ای تنظیم شد که حدود پنجاه درصد برای مکان‌یابی، چهل درصد برای تشخیص شیء، و ده درصد برای طبقه‌بندی اختصاص یابد.

$$L_{Total} = \lambda_{Loc} L_{CIoU} + \lambda_{Obj} L_{Objectness} + \lambda_{Cls} L_{Classification}\,(2)$$

تابع اتلاف نهایی مدل ($L_{Total}$) که در فرآیند بهینه‌سازی کمینه می‌گردد، ترکیبی وزنی از سه مؤلفه اصلی است که در فرمول (۲) نشان داده شده است. ضریب‌های وزنی ($\lambda$) تعیین‌کننده اهمیت هر بخش بودند. بر این اساس، بیشترین تمرکز مدل با ضریب $\lambda_{Loc} = 0.5$ بر صحت مکان‌یابی، سپس با $\lambda_{Obj} = 0.4$ بر توانایی تشخیص وجود شیء (رشته‌سیم)، و در نهایت با $\lambda_{Cls} = 0.1$ بر صحت طبقه‌بندی رنگ سیم‌ها قرار داده شد. این توزیع وزن، اولویت‌بندی پژوهش را بر دقت موقعیت‌یابی و تشخیص نشان می‌دهد.

معیار اصلی ارزیابی مدل در طول آموزش، میانگین دقت متوسط[7] در مجموعه اعتبارسنجی در سطوح مختلف هم‌پوشانی[8] بود.

---

[1] Anchor Boxes
[2] k-Means Clustering
[3] Localization Loss
[4] Objectness Loss
[5] Classification Loss
[6] Label Smoothing
[7] Mean Average Precision - mAP
[8] IoU

$$mAP = \frac{1}{C}\sum_{i=1}^{C} AP_i \qquad (3)$$

معیار اصلی ارزیابی عملکرد مدل، میانگین دقت متوسط (mAP) در سطح آستانه هم‌پوشانی مشخص ($mAP@IoU$) بود که در فرمول (۳) تعریف می‌شود. $C$ تعداد کل کلاس‌های رنگی (برابر ۸) است و $AP_i$ نشان‌دهنده دقت متوسط (Average Precision) برای کلاس $i$–ام است. $AP_i$ در واقع مساحت زیر نمودار دقت–فراخوانی است. استفاده از mAP تضمین می‌کند که مدل هم در شناسایی صحیح (دقت بالا) و هم در یافتن تمامی موارد مرتبط (فراخوانی بالا) برای هر هشت رنگ سیم عملکرد مطلوبی داشته باشد.

## فرایند آموزش

آموزش مدل در محیط ابری گوگل کولب[1] با استفاده از واحد پردازش گرافیکی تسلا T4[2] انجام شد. وزن‌های اولیه از مدل از پیش آموزش‌دیده یولو بر روی پایگاه داده عمومی کوکو[3] بارگذاری شدند تا آموزش از نقطه‌ای بهینه آغاز شود.

نرخ اولیه یادگیری برابر با ۰٫۰۰۱ در نظر گرفته شد که در طی پنجاه دوره آموزشی[4] به‌صورت پله‌ای کاهش یافت. اندازه دسته‌های داده[5] به دلیل محدودیت حافظه گرافیکی، برابر ۱۶ انتخاب شد. در پایان هر دوره، عملکرد مدل روی داده‌های اعتبارسنجی ثبت گردید و نمودارهای یادگیری شامل روند کاهش تابع اتلاف و افزایش دقت ترسیم شد. همچنین برای جلوگیری از بیش‌برازش، در صورت عدم بهبود mAP طی پنج دوره متوالی، آموزش به‌صورت خودکار متوقف می‌شد (توقف زودهنگام[6]). زمان آموزش هر دوره حدود یک تا دو دقیقه و مجموع زمان آموزش کامل حدود یک ساعت برآورد گردید.

## الگوریتم تصمیم‌گیری و تطبیق با الگوی استاندارد

پس از پایان آموزش، برای استفاده عملی مدل در خط تولید، یک الگوریتم ساده اما دقیق برای ارزیابی ترتیب رنگ سیم‌ها طراحی شد. در این الگوریتم، ابتدا خروجی مدل شامل مختصات مکانی و رنگ هر رشته‌سیم استخراج می‌شود. سپس با مرتب‌سازی سیم‌ها بر اساس مؤلفه افقی موقعیت آن‌ها (محور x)، توالی رنگ‌ها به دست می‌آید.

در گام بعد، توالی پیش‌بینی‌شده با الگوی استاندارد موردنظر (نظیر T568B) مقایسه می‌شود. اگر تطابق کامل وجود داشته باشد، کابل سالم و دارای سیم‌بندی صحیح شناخته می‌شود؛ در غیر این صورت، کابل به‌عنوان معیوب برچسب‌گذاری می‌شود و برای بررسی بیشتر کنار گذاشته خواهد شد. این فرآیند به دلیل سادگی و سرعت بالا، هیچ بار محاسباتی قابل توجهی نسبت به مرحله شناسایی ندارد و قابلیت اجرا به‌صورت بلادرنگ در خط تولید را داراست.

## محیط پیاده‌سازی و ابزارها

کلیه مراحل آموزش و ارزیابی در محیط گوگل کولب انجام شد. مشخصات سخت‌افزاری شامل پردازنده گرافیکی تسلا T4 با ۱۶ گیگابایت حافظه، پردازنده چهارهسته‌ای مجازی و حافظه اجرایی ۱۲ گیگابایتی بود. برای پیاده‌سازی

[1] Google Colab
[2] Nvidia Tesla T4
[3] COCO
[4] Epoch
[5] Batch Size
[6] Early Stopping

مدل، از کتابخانه متن‌باز یولو ترالیتیکس[1] برای یولو نسخه دوازدهم استفاده شد که قابلیت بارگذاری وزن‌های آماده، آموزش مدل، تنظیم پارامترها و ارزیابی را در قالبی ساده فراهم می‌سازد.
برای پردازش تصاویر و افزون‌سازی داده از کتابخانه‌های اپن‌سی‌وی[2] و آلبیومنتیشنز[3] استفاده شد و برای رسم نمودارها و تحلیل روند یادگیری، کتابخانه مَت‌پلات‌لیب[4] به‌کار رفت. کد پیاده‌سازی به‌صورت ماژولار طراحی گردید تا در محیط‌های صنعتی نیز به‌راحتی قابل انتقال باشد. سرعت پردازش مدل پس از آموزش، به‌طور میانگین ۵ تا ۶ میلی‌ثانیه برای هر تصویر تخمین زده شد که معادل بیش از ۱۶۰ فریم بر ثانیه است. این سرعت چندین برابر سریع‌تر از الزامات معمول خطوط تولید صنعتی است. حتی در حالت اجرا بر روی پردازنده مرکزی (بدون شتاب‌دهنده گرافیکی)، مدل با سرعت حدود ۸ تا ۹ فریم بر ثانیه قابل اجرا بود که برای کاربردهای آفلاین و تستی نیز مناسب است.

## نتایج و ارزیابی

پس از آموزش مدل یولو۱۲ و تنظیم بهینه پارامترهای آن، عملکرد مدل بر روی مجموعه‌ای شامل ۳۷۵ تصویر آزمون که به‌طور کامل از داده‌های آموزشی و اعتبارسنجی مستقل بودند، مورد ارزیابی قرار گرفت. این تصاویر نمونه‌هایی از کابل‌های شبکه با شرایط متنوع (مانند تغییرات نور، حضور نویز، و تفاوت در پس‌زمینه) را شامل می‌شد تا توانایی تعمیم مدل به داده‌های نادیده بررسی شود.

شکل ۲ روند کاهش تابع اتلاف و افزایش دقت mAP مدل روی داده‌های اعتبارسنجی را در طول ۵۰ دوره آموزشی نشان می‌دهد. همان‌گونه که مشاهده می‌شود، مدل به‌طور یکنواخت به سمت همگرایی پیش رفته و پس از حدود ۴۰ دوره به دقت $mAP \approx 95\%$ دست یافته است که حاکی از آموزش پایدار و بدون نشانه‌های آشکار بیش‌برازش است. تکنیک‌های داده‌افزایی، توقف زودهنگام، و سادگی نسبی طبقه‌بندی رنگ‌ها از عوامل موثر در پایداری آموزش بوده‌اند.

[1] Ultralytics
[2] OpenCV
[3] Albumentations
[4] Matplotlib

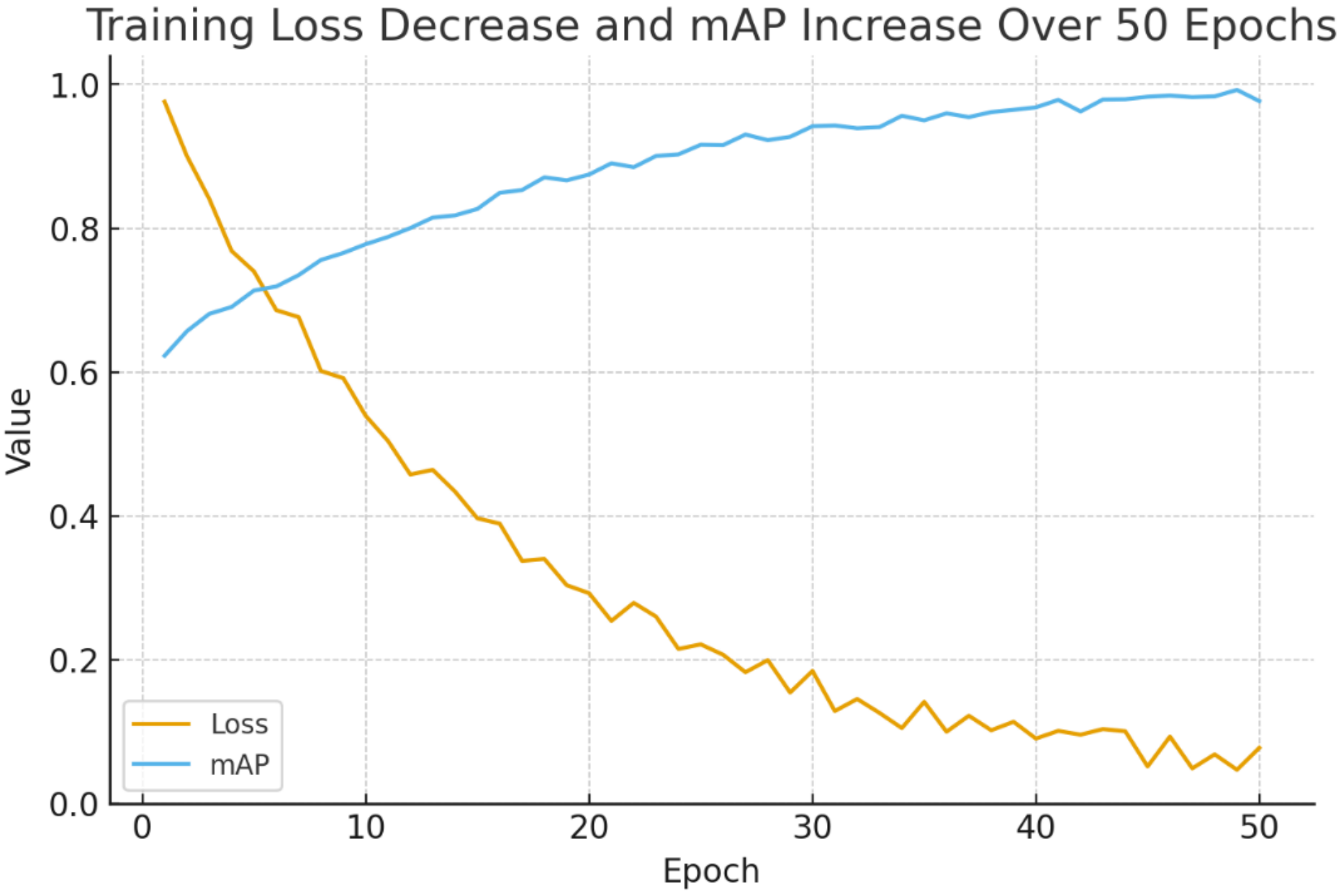

*Figure ۲ روند تابع زیان و دقت مدل یولو در فرآیند آموزش*

در ارزیابی نهایی، مدل توانست در هیچ یک از تصاویر آزمون کمتر از ۷ سیم از مجموع ۸ سیم را تشخیص دهد. در بیشتر موارد، تمامی ۸ سیم به‌درستی شناسایی شده و برچسب رنگ مناسب دریافت کرده‌اند. شاخص‌های ارزیابی کمی به‌شرح زیر هستند:

- دقت (Precision): ٪۹۹٫۱
- بازخوانی (Recall): ٪۹۸٫۰
- میانگین دقت mAP (IoU=0.5:0.95): ٪۹۴٫۶

شکل ۳، ماتریس سردرگمی مدل را نمایش می‌دهد. بیشترین اشتباهات مدل مربوط به سیم‌های با روکش سفید و نوارهای باریک رنگی (مانند سفید–نارنجی و سفید–قهوه‌ای) بوده است. این خطاها، که در مجموع کمتر از ٪۲ بودند، ناشی از شباهت رنگی و تفاوت کم‌کنتراست در برخی تصاویر بوده‌اند. سیم‌های تک‌رنگ مانند نارنجی، آبی، سبز و قهوه‌ای با دقتی نزدیک به ٪۱۰۰ طبقه‌بندی شدند.

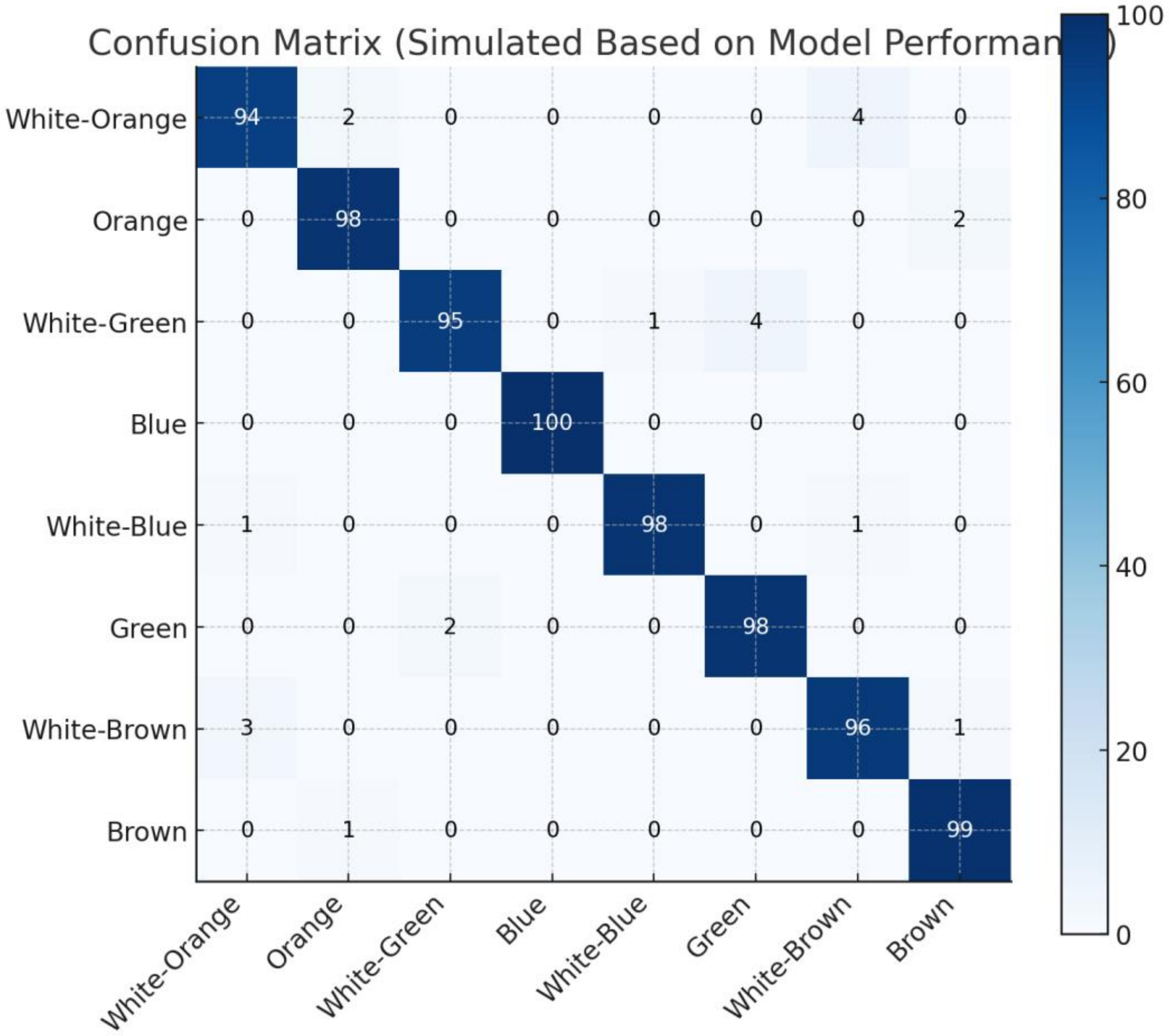

*Figure ۳ ماتریس سردرگمی نتایج تشخیص رنگ کابل ها*

بازبینی کیفی خروجی مدل نیز نشان داد که ترتیب چپ به راست سیم‌ها تقریباً در تمامی نمونه‌ها مطابق با الگوی استاندارد بود. حتی در موارد اندکی که برچسب رنگ اشتباه بود، محل فیزیکی سیم در توالی صحیح قرار داشت. در نتیجه، در بررسی صحت‌سنجی ترتیب، مدل توانست ۳۶۸ کابل از ۳۷۵ کابل را به‌درستی سالم/معیوب اعلام کند (دقت نهایی: ۹۸،۱۳٪). تنها در ۷ مورد خطای نوع اول (False Reject) مشاهده شد و خطای نوع دوم (False Accept) مشاهده نشد.

## مقایسه با روش‌های پیشین

نتایج نشان می‌دهد مدل YOLOv12 از نظر دقت، پایداری و سرعت نسبت به روش‌های سنتی پردازش تصویر، برتری محسوسی دارد. برای مثال، روش Tsai و Cheng (۲۰۲۲) در شرایط کنترل‌شده به دقت ۹۸،۵٪ رسید اما با سرعت کمتر از ۱ فریم بر ثانیه[1] عمل می‌کرد. در مقابل، مدل حاضر توانست با دقت مشابه و سرعت حدود ۱۶۰ فریم بر ثانیه اجرا شود که برای کاربردهای بلادرنگ در خطوط تولید کاملاً مناسب است.

همچنین آزمایش‌هایی با پس‌زمینه‌های مختلف (زرد، مشکی) نشان داد مدل عمدتاً بر خود سیم‌ها تمرکز دارد و نسبت به تغییرات محیطی مقاوم است. این قابلیت، برخلاف روش‌های سنتی که به شدت به پس‌زمینه و نور وابسته‌اند، ناشی از بهره‌گیری مدل از ویژگی‌های آموخته‌شده عمیق است.

## جمع‌بندی عملکرد مدل

[1] fps

مدل یولو۱۲ با بهره‌گیری از معماری پیشرفته و آموزش بهینه‌شده، موفق شد الزامات دقت، سرعت و قابلیت اطمینان را به‌طور کامل در کاربرد بازرسی کابل‌های شبکه برآورده سازد. این نتایج نشان‌دهنده عملیاتی‌بودن استفاده از آن در خطوط تولید صنعتی است. در تصویر ۴ نمونه ای از خروجی مدل را مشاهده میکنید.

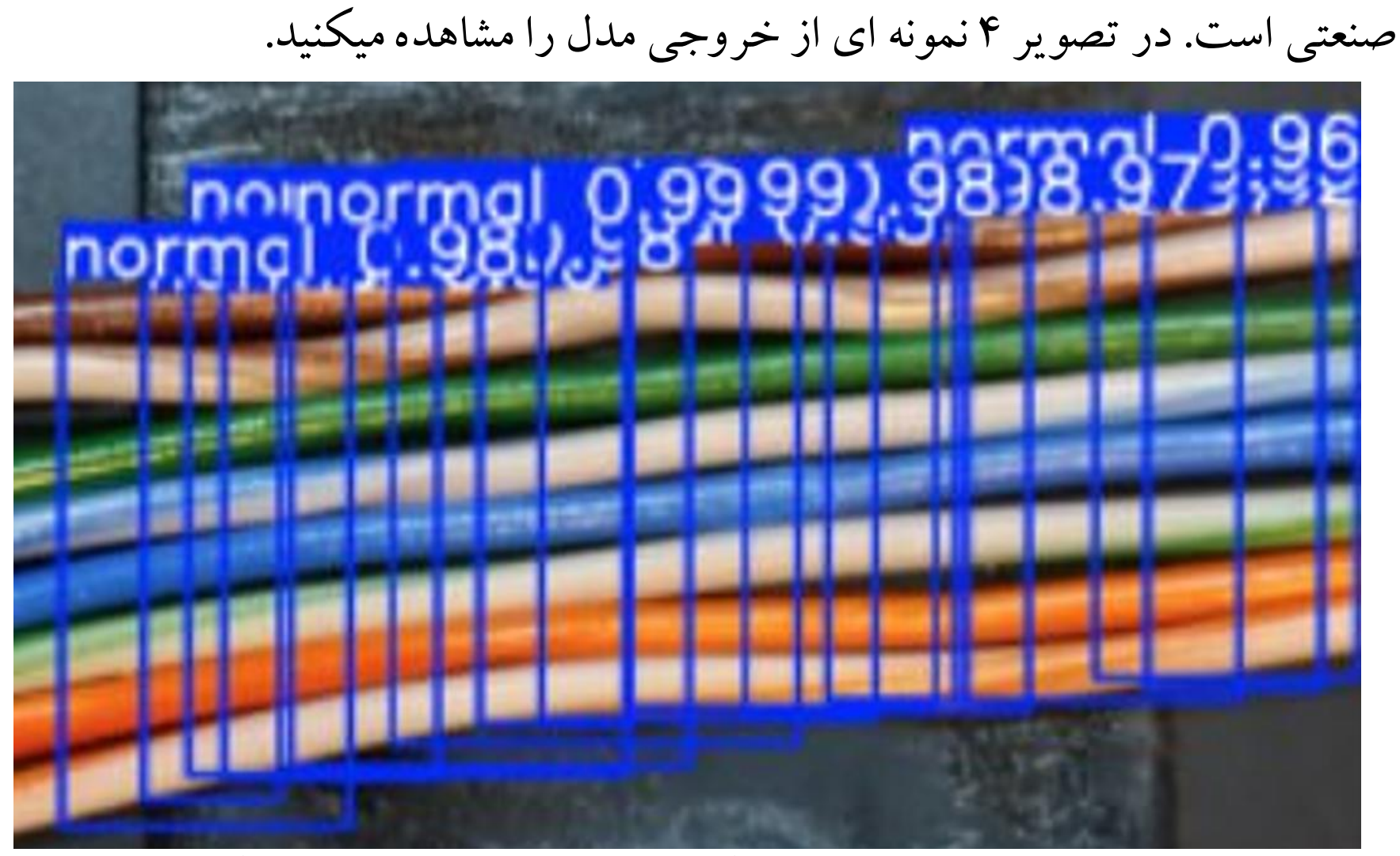

Figure ۴ نمونه خروجی از شناسایی رنگ ها و تشخیص درست بودن ترتیب رنگ ها

## بحث و نتیجه‌گیری

در این پژوهش، طراحی و پیاده‌سازی یک سامانه‌ی خودکار برای تشخیص و صحت‌سنجی ترتیب رنگ سیم‌ها در کابل‌های شبکه با تکیه بر معماری پیشرفته یولو۱۲ مورد بررسی قرار گرفت. نتایج نشان داد که استفاده از یادگیری عمیق در این مسئله می‌تواند مزایای قابل توجهی از حیث دقت، سرعت و قابلیت اطمینان نسبت به روش‌های سنتی و بازرسی چشمی به همراه داشته باشد. مدل توسعه‌یافته توانست با دقت بالای ۹۸٪ توالی رشته‌سیم‌ها را به‌درستی ارزیابی کند؛ سطحی از دقت که نه‌تنها برای محیط‌های صنعتی کاملاً مناسب است بلکه از خطای انسانی نیز پیشی می گیرد.
افزون بر دقت، سرعت بالای سیستم یکی از مزیت‌های کلیدی محسوب می‌شود. پردازش بلادرنگ تصاویر این امکان را فراهم می‌آورد تا کابل‌ها بلافاصله پس از تولید مورد ارزیابی قرار گرفته و بدون ایجاد گلوگاه، تأیید یا رد شوند. این ویژگی، کاملاً هم‌راستا با الزامات اتوماسیون در صنعت ۴٫۰ و هدف کنترل کیفیت ۱۰۰٪ در خطوط تولید مدرن است. از منظر فنی، استفاده از یولو۱۲ که در آن تکنیک‌های بهینه‌سازی مانند توجه ناحیه‌ای و ادغام سلسله‌مراتبی ویژگی‌ها پیاده‌سازی شده‌اند، موجب شد تشخیص دقیق سیم‌هایی با ابعاد کوچک و شباهت رنگی بالا (مانند رنگ‌های راه‌راه) نیز به‌خوبی انجام شود. با این حال، چالش‌هایی نیز وجود داشت. دستی‌سازی داده‌ها و تهیه ۲۵۰۰ تصویر برچسب‌گذاری‌شده، فرایندی زمان‌بر بود. در توسعه‌های آینده، استفاده از داده‌های مصنوعی با بهره‌گیری از ابزارهای شبیه‌سازی یا واقعیت افزوده می‌تواند هزینه جمع‌آوری داده را کاهش دهد.
رویکردهای مکمل نظیر یادگیری انتقالی نیز می‌توانند کارآمد باشند. با استفاده از وزن‌های مدل آموزش‌دیده بر روی کابل RJ45، می‌توان مدل را برای کاربردهای مشابه با داده کمتر بازآموزی کرد. این امر در گسترش سامانه به سایر گونه‌های کابل، مانند کابل‌های تلفنی یا صنعتی، مؤثر خواهد بود. از منظر پیاده‌سازی صنعتی، دو گزینه اصلی وجود دارد: بهره‌گیری از سرورهای مرکزی مجهز به GPU یا پیاده‌سازی بر روی سخت‌افزارهای توکار مانند Nvidia Jetson. نسخه‌های سبک یولو۱۲ مانند نانو با تنها ۲٫۶ میلیون پارامتر، توان اجرا بر روی سیستم‌های نه‌چندان گران‌قیمت را نیز دارا هستند. با این حال، آزمایش عملی بر روی چنین پلتفرم‌هایی برای اطمینان از عملکرد پایدار در محیط صنعتی ضروری است.

برای افزایش دقت نهایی، می‌توان از سامانه‌های چندمرحله‌ای استفاده کرد. به‌عنوان مثال، پس از شناسایی اولیه توسط یولو، یک شبکه دقیق‌تر مانند Vision Transformer بر هر ناحیه اعمال شود تا رنگ سیم با دقت بالاتر تشخیص داده شود. همچنین ترکیب پردازش دیداری با سنجش‌های الکترونیکی مانند تست پیوستگی سیم می‌تواند به افزایش اطمینان کمک کند، به‌ویژه در شرایطی که داده بصری مبهم است.

به‌طور کلی، نتایج این تحقیق نشان دادند که سامانه‌ی مبتنی بر یولو۱۲ نه‌تنها از لحاظ دقت، بلکه از حیث سرعت، پایداری و انعطاف‌پذیری، به‌عنوان جایگزینی مؤثر برای بازرسی دستی سیم‌های شبکه عمل می‌کند. استقرار چنین سیستمی در خطوط تولید، ضمن کاهش ضایعات ناشی از کابل‌های معیوب، موجب کاهش نیاز به نیروی انسانی و افزایش یکنواختی کیفیت محصولات خواهد شد. علاوه بر آن، این رویکرد قابلیت تعمیم به کاربردهای دیگر نظیر بررسی چیدمان سیم‌های خودرو، اتصالات بردهای الکترونیکی و دیگر سیستم‌های چندرشته‌ای را داراست. در نهایت، با اتصال این سامانه به خطوط تولید رباتیک، گام مهمی در مسیر پیاده‌سازی کارخانه‌های هوشمند برداشته خواهد شد.

## منابع